\pdfoutput=1
\documentclass[letterpaper, 10pt, conference]{ieeeconf}

\usepackage{graphics}
\usepackage{epsfig}

\usepackage{cite}
\usepackage[dvipsnames, svgnames]{xcolor}
\usepackage{xurl}
\usepackage{hyperref}
\hypersetup{
    colorlinks,
    linkcolor={red!50!black},
    citecolor={blue!50!black},
    urlcolor={blue!80!black}
}
\usepackage{bm}
\usepackage{breqn}
\usepackage{amssymb}
\usepackage{tabularx}
\usepackage{booktabs}
\usepackage{dcolumn}
\newcolumntype{d}[1]{D..{#1}}
\newcolumntype{C}{>{\centering\arraybackslash}X}
\usepackage{soul}
\usepackage[linesnumbered,ruled,vlined]{algorithm2e}

\usepackage{amsthm}

\newcommand{\reals}{\mathbb{R}}

\newcommand{\R}{\reals}

\newcommand{\Bcal}{\mathcal{B}}
\newcommand{\Ccal}{\mathcal{C}}

\newcommand{\Rcal}{\mathcal{R}}
\newcommand{\Scal}{\mathcal{S}}

\newcommand{\Ucal}{\mathcal{U}}

\newcommand{\Xcal}{\mathcal{X}}

\newcommand{\eqn}[1]{\begin{align} #1 \end{align}}

\theoremstyle{plain}

\theoremstyle{definition}

\theoremstyle{remark}

\definecolor{yellow}{cmyk}{0.0,0.10,0.95,0.0}
\definecolor{pred}{cmyk}{0,0.8,0.70,0.0}
\definecolor{bluedefined}{cmyk}{0.46, 0.10, 0, 0.0}

\IEEEoverridecommandlockouts  
\usepackage[english]{babel}
\usepackage{footnotehyper}

\usepackage{hyperref,tablefootnote,footnotehyper}
\usepackage{amsmath,booktabs,mwe,pdflscape}
\hypersetup{colorlinks}

\usepackage{leftindex} 
\usepackage{graphicx} 
\usepackage{amsmath}
\usepackage{amssymb}
\usepackage{amsthm}
\usepackage{amsfonts}
\usepackage{float}
\usepackage{mathrsfs}
\usepackage{booktabs}
\usepackage{array}
\usepackage[nodisplayskipstretch]{setspace}
\makeatletter
\def\BState{\State\hskip-\ALG@thistlm}
\makeatother
\usepackage[capitalize,nameinlink,noabbrev]{cleveref} 
\theoremstyle{plain}

\theoremstyle{definition}

\newcolumntype{M}[1]{>{\centering\arraybackslash}m{#1}}
\usepackage{pifont}
\usepackage{multirow}
\usepackage{tabularray}
\usepackage{cite}
\usepackage{algpseudocode}

\newcommand{\RNum}[1]{\uppercase\expandafter{\romannumeral #1\relax}}
\usepackage{bigints}

\newcommand{\gatekeeper}{\textbf{\texttt{gatekeeper}}}

\newcommand{\eware}{\textbf{\texttt{eware}}}
\newcommand{\gware}{\textbf{\texttt{gware}}}
\newcommand{\mesch}{\textbf{\texttt{meSch}}}
\newcommand{\reroot}{\textbf{\texttt{ReRoot}}}
\algrenewcommand\textproc{}
\newcommand{\resilient}

\usepackage[font=small,skip=10pt]{caption}

\algrenewcommand\algorithmicrequire{\textbf{Input:}}
\algrenewcommand\algorithmicensure{\textbf{Output:}}

\hypersetup{
  colorlinks,
  citecolor=teal, 
  linkcolor=teal,
  urlcolor=purple}

\author{Kaleb Ben Naveed$^{1,\dag}$, Devansh R. Agrawal$^{1,2, \dag}$, Daniel M. Cherenson$^{1,\dag}$, Haejoon Lee$^{1}$, Alia Gilbert$^{1}$,\\ Hardik Parwana$^{1}$, Vishnu S. Chipade$^{3}$, William Bentz$^{4}$, and Dimitra Panagou$^{1,2}$%
\thanks{$\dag$Equal Contribution}
\thanks{$^{1}$Department of Robotics, University of Michigan, Ann Arbor, MI, 48109 USA. $^{2}$Department of Aerospace Engineering, University of Michigan, Ann Arbor, MI, 48109 USA. $^{3}$Independent Researcher. $^{4}$NASA Goddard Space Flight Center. Corresponding authors: {\tt\small \{kbnaveed, dpanagou\}@umich.edu}}
}
\title{\LARGE \bf
Enabling Safety for Aerial Robots: Planning and Control Architectures
} 

\def\arraystretch{1.2}

\newcommand\footnoteref[1]{\protected@xdef\@thefnmark{\ref{#1}}\@footnotemark}

\begin{document}

\maketitle
\thispagestyle{empty}
\pagestyle{empty}

\begin{abstract}
Ensuring safe autonomy is crucial for deploying aerial robots in real-world applications. However, safety is a multifaceted challenge that must be addressed from multiple perspectives, including navigation in dynamic environments, operation under resource constraints, and robustness against adversarial attacks and uncertainties. In this paper, we present the authors' recent work that tackles some of these challenges and highlights key aspects that must be considered to enhance the safety and performance of autonomous aerial systems. All presented approaches are validated through hardware experiments.
\end{abstract}



\section{Introduction}
Aerial robots are employed in various tasks, including search and rescue, environmental monitoring, infrastructure inspection, and delivery services. Ensuring safe autonomy is crucial for their reliable deployment in real-world applications. However, achieving safety in autonomous systems is highly challenging due to several factors. One key challenge is that safety is a multifaceted issue, requiring consideration from multiple perspectives to ensure reliable operation. For instance, aerial robots must navigate unknown and dynamic environments, operate within energy and budget constraints, and remain robust against adversarial attacks and disturbances. At the same time, safety must be upheld without overly compromising mission objectives.
In this paper, we present the authors' recent works that address these challenges to improve aerial robot capabilities. 
We discuss three aspects of safety: contingency planning, adversarial resilience, and robustness under uncertainty and disturbances.

 In contingency planning, safety is maintained by switching to a backup trajectory whenever an unsafe condition is detected~\cite{Dev_TRO_2024_gatekeeper}. At each planning iteration, a candidate trajectory is generated, consisting of a segment of the nominal mission-optimized trajectory followed by the backup. Before fully executing the nominal portion of the last committed trajectory, a new candidate is generated. If deemed safe, it is committed for future tracking; otherwise, the robot continues following the last committed trajectory until it reaches a safe set. We also describe how this approach can be applied to energy-aware~\cite{Kaleb_ICRA_2024_Eclares, Kaleb_IROS_2025_meSch} and resource-constrained planning.

Ensuring safety in the presence of adversarial influence poses a greater challenge. Generally, adversarial influences can compromise a robot’s safety in two ways: by corrupting informational security through faulty data injection or by threatening physical safety through direct physical attacks. Adversarial resilience addresses this challenge by leveraging information redundancy to filter out corrupted data or by actively detecting and mitigating adversarial threats to minimize their impacts. In this work, we examine how adversarial resilience can be used to ensure safety of multi-robot systems~\cite{Haejoon_ICRA_2024, Vishnu_Frontiers_2021}.

In safe planning under uncertainty, safety is maintained by adapting control strategies to account for unknown disturbances and imperfect information. Uncertainty arises from various sources, including sensor noise, disturbances in system dynamics, and imperfect knowledge of the environment. To address these challenges, we present methods that enable robots to operate safely despite these uncertainties. Specifically, we discuss  online adaptation of control policies for uncertain nonlinear systems using model-based reinforcement learning~\cite{Hardik_TCST_2025_foresee}, robust observer-controller synthesis~\cite{Dev_CSL_2023}, and resilient coordination strategies for multi-robot systems~\cite{Alia_ACC_2022}. These approaches help aerial robots quantify and mitigate uncertainty, adapt in real time, and ensure safe operation in dynamic environments.

Finally, we introduce DevQuad, an agile, in-house-developed quadrotor used for a subset of experiments.

\section{Safety Through Contingency Planning}
\begin{table*}[h]
    \scriptsize
    \centering
    \renewcommand{\arraystretch}{1.3} 
    \begin{tabular}{>{\arraybackslash}p{3.9cm}>{\arraybackslash}p{6.9cm}>{\arraybackslash}p{2.0cm}>{\arraybackslash}p{1.5cm}>{\arraybackslash}p{1.2cm}}
        \hline
        \textbf{Method Demonstrated} & \textbf{Application} & \textbf{Platform Used}& \textbf{Section} & \textbf{Figure}\\ 
        \hline
        \gatekeeper{}~\cite{Dev_TRO_2024_gatekeeper} & Safe navigation and exploration in unknown environments & DevQuad $\times 1$ &  \cref{gatekeeper} & \cref{fig:snapshots}(a)\\ 
        \hline
        \eware{}~\cite{Kaleb_ICRA_2024_Eclares} & Energy-aware ergodic search & DevQuad $\times 1$ &  \cref{eware_meSch} & \cref{fig:snapshots}(b)\\ 
        \hline
        \mesch{}~\cite{Kaleb_IROS_2025_meSch} & Multi-agent energy-aware ergodic search with shared charging station & DevQuad $\times 3$ &  \cref{eware_meSch} & \cref{fig:snapshots}(c)\\ 
        \hline
        Resilience-aware controller~\cite{Haejoon_ICRA_2024} & Resilient leader-follower network formation & Crazyflie 2.0 $\times 8$ &  \cref{joon_resilience_aware_Control} & \cref{fig:snapshots}(d)\\
        \hline
        StringNet~\cite{Vishnu_Frontiers_2021} & Safety against adversarial agents & F330  $\times 4$ &  \cref{StringNet} & \cref{fig:snapshots}(e)\\
        \hline
        FORESEE~\cite{Hardik_TCST_2025_foresee} & Adaptive trajectory tracking with collision avoidance & DevQuad $\times 1$ &  \cref{forsee} & \cref{fig:snapshots}(f)\\
        \hline
        Robust observer-controller~\cite{Dev_CSL_2023} & Robust trajectory tracking under observer uncertainty & Crazyflie 2.0 $\times 1$ &  \cref{observer_controller} & \cref{fig:snapshots}(g)\\
        \hline
         Formation control with humans~\cite{Alia_ACC_2022} & Robust leader follower formation controls for human-robot scenarios & F330  $\times 4$ &  \cref{robust_leader_follower} & \cref{fig:snapshots}(h)\\
        \hline
         Dynamic energy-aware coverage~\cite{William_Autonomous_Robots_2018} & Complete 3-D dynamic coverage in energy-constrained networks & Hummingbird  $\times 3$ &  \cref{eware_meSch} & \cref{fig:snapshots}(i,j,k)\\
        \hline
    \end{tabular}
    \caption{Summary of hardware experiments}
    \label{tab:Experiments}
\end{table*}
\begin{figure*}[t]
  \centering
  \includegraphics[width=2.0\columnwidth]{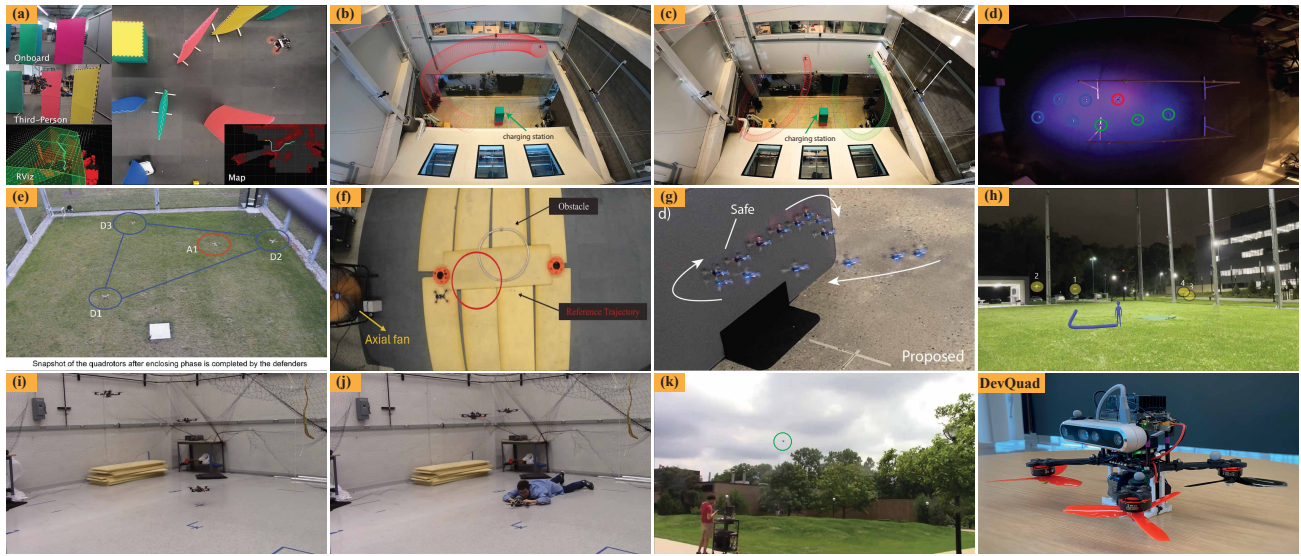}
  \caption{Experiment snapshots}
  \vspace{-17pt}
  \label{fig:snapshots}
\end{figure*}
We first mathematically formalize the notion of safety. Consider a dynamical system:
\eqn{
\label{sys_dyn}
\Dot{x} = f(x, u),
}
where $x \in \Xcal \subset \R^n$ is the robot state and $u \in \Ucal \subset \R^m$ is the control input. The system is subject to constraints that define a constrained, or safe set $\mathcal S$ given as:
\eqn{
\Scal = \{ x \; | \; h_1(x) \geq 0, h_2(x) \geq 0, \cdots \},
}
where $h_i:\mathcal X \rightarrow \mathbb R$, $i\in \{1,2,
\dots\}$ are functions capturing constraints such as avoiding obstacles, maintaining minimum energy levels, etc. The goal of the autonomy stack is to ensure that the closed-loop system satisfies:
\eqn{
\label{safety_req}
x(t) \in \Scal \quad \forall t \geq t_0.
}

\subsection{\gatekeeper}\label{gatekeeper}
\gatekeeper{} introduced in~\cite{Dev_TRO_2024_gatekeeper} enforces \eqref{safety_req} by guaranteeing that there exists a safe trajectory $p : [t_0, \infty) \rightarrow \Xcal$ from the current state: $\exists \ p(t) \in \Scal \quad \forall t \geq t_0, \quad p(t_0) = x(t_0)$.

\gatekeeper{} is an iterative process that ensures that only safe trajectories are allowed to be executed by the aerial robot at each iteration $k$ of the autonomy stack. To do this, it uses the concept of a perceived safe set $\Bcal_k$, and of a backup safe set $\Ccal_k$. The perceived safe set $\Bcal_k$ is constructed using the sensory information available up to time $t_k$, and is possibly time-varying. Similarly, the backup safe set $\Ccal_k$ (also potentially time-varying) represents the set where the robot should terminate its trajectory in case a violation of safety is predicted. More specifically, at each iteration $k$, \gatekeeper{} constructs a candidate trajectory (defined later) using newly available information, checks whether the candidate trajectory is valid, and if so, replaces the old committed trajectory with the new candidate trajectory. The candidate trajectory is constructed as a concatenation ``stitching'' of the nominal mission-optimized trajectory and of the backup trajectory, which by construction terminates at the backup set, which is a robustly controlled-invariant set. The candidate trajectory is considered valid if it lies strictly within the perceived safe set. Thus, if a candidate trajectory is valid, it can be safely tracked for all $t \geq t_k$, otherwise the committed trajectory from the prior iteration is used and tracked by the controller. Since the committed trajectory is updated with a candidate trajectory only if the latter is valid, it follows that the committed trajectory can always be safely tracked.

\subsection{Energy-Aware Planning: Introducing \eware{} and \mesch{}}\label{eware_meSch}
The Energy-Aware Filter (\eware{}) introduced in~\cite{Kaleb_ICRA_2024_Eclares} is an application of \gatekeeper{}. \eware{} was developed for persistent missions, which require robots to return to a base, e.g., their charging station, when needed. At each iteration of the algorithm, candidate trajectories are generated that follow a portion of the nominal mission trajectory before transitioning to the back-to-base (b2b) trajectory leading to the charging station. In \eware{}, the perceived safe set includes all system states with non-zero energy, while the backup safe set consists of all robot states within a certain radius of the charging station.

\mesch{} extends \eware{} to the multi-agent case, enabling persistent missions involving multiple robots. An additional constraint was introduced, requiring all robots to share a single charging station, which could be mobile. To manage this, a constraint was enforced to ensure that robots visit the charging station with sufficient time gaps. The \gware{} module maintains the minimum time gap between charging sessions by constructing gap flags and rescheduling robots to arrive earlier when conflicts arise, ensuring the gap condition is always met. Once all gap constraints are satisfied, \eware{} verifies whether each robot has enough energy to continue its mission, ensuring the minimum energy condition is upheld.  Moreover, we have demonstrated that \eware{} and \mesch{} can serve as low-level filters in adaptive exploration missions, particularly in ergodic search~\cite{Kaleb_ICRA_2024_Eclares, Kaleb_IROS_2025_meSch}. 

Conceptually similar is our earlier line of work on 3D energy-aware dynamic coverage \cite{William_Autonomous_Robots_2018}. Our approach uses a novel domain partitioning strategy that directs individual agents to explore within concentric hemispherical shells around a centralized charging station. This design ensures that flight paths naturally lead agents back to the charging station as their batteries deplete, enabling efficient coverage and recharge cycles.

\subsection{Budget-Constrained Safe Planning}
We extend \gatekeeper{} to provide a general framework for enforcing a budget constraint on an expendable resource, denoted $b(t) \le B$, which is consumed during the mission and can be renewed when the robot enters designated regions $\Rcal$~\cite{cherenson2025autonomy}. Previously, \eware{} focused specifically on energy constraints, where energy is renewed at charging stations. Here, we generalize this concept to other resource limitations, such as accumulated localization error in visual odometry, which can be renewed near visual landmarks. If the renewal set $\Rcal$ is assumed to be a backup set, \gatekeeper{} guarantees that both the safety and budget constraints are satisfied for all time. Additionally, we introduce a backup trajectory planner, \reroot{} (Reverse Rooted Forest), a sampling-based method built on RRT*~\cite{karaman2011sampling}, to generate minimal budget paths from any initial condition to the renewal region $\Rcal$. Multiple rooted trees are grown from multiple renewal regions and extend into the perceived safe set, allowing for a backup trajectory to be generated quickly by connecting to a nearby node and following the tree to the root.  


\section{Safety in the Presence of Adversaries}

\subsection{Resilience-Aware Control}\label{joon_resilience_aware_Control}
The leader-follower consensus problem in a network enables robots called followers to converge to the reference state value propagated by robots known as leaders~\cite{dimar2008}. This allows the robots to coordinate their actions to achieve the common objective. However, in the presence of adversaries that actively propagate false information in the network, consensus may be disrupted, potentially putting robots' safety at risk. There has been a large body of literature studying the \textit{Weighted Mean Subsequence Reduced} W-MSR algorithms~\cite{wmsr_2013,James_TAC_2020} to ensure leader-follower consensus despite adversarial influences.

Although the W-MSR algorithms are simple and only require local information, many of them rely on network resilience properties. As a result, these network properties have been well-studied~\cite{Joon_CDC_2024, James_TAC_2020}, but in general they depend on global network states and are hard to represent analytically, making their implementation in multi-robot systems challenging. 
In~\cite{Haejoon_ICRA_2024}, we have designed a resilience-aware controller that ensures multi-robot network maintains strong $r$-robustness~\cite{James_TAC_2020}, the network property for resilient leader-follower consensus, without fixed network topologies during its operation. In essence, we have constructed a constraint set~\eqref{safety_req} to represent a set of collective states of robots whose network maintains strong $r$-robustness above a given threshold for all time.

\subsection{Aerial Swarm Defense by StringNet Herding}\label{StringNet}
The work introduced in~\cite{Vishnu_Frontiers_2021} studied the problem of defending a protected area from multiple aerial attackers or intruders using a swarm of aerial defenders. The challenge lies in effectively coordinating the defenders to herd the attackers away while accounting for real-time decision-making constraints and computational complexity. Traditional interception-based methods may not scale well with a large number of attackers, making efficient swarm-based strategies necessary.

To address this, we propose a StringNet herding approach, where defenders dynamically form a net-like structure to guide attackers toward safe areas. The method consists of two phases: a gathering phase, where defenders determine an optimal interception point using a Mixed Integer Quadratic Programming (MIQP) formulation, and a herding phase, where they form an open-line formation to steer attackers away. Since MIQP scales with the number of robots and becomes computationally expensive for large swarms, we propose an approximate solution to improve feasibility checks, ensuring scalability and computational efficiency. Additionally, we provide conditions under which defenders can successfully herd all attackers.

\section{Safety Under Uncertainty}

\subsection{Prediction with Expansion-Compression Unscented Transform for Online Policy Optimization}\label{forsee}
In the real world, perfect knowledge of system models is seldom available, leading to degradation in performance and loss of safety guarantees for controllers and planners designed for nominal dynamics. Consequently, probabilistic dynamics with generic state-dependent uncertainties
\begin{align}
    \dot x \sim D(x, u) 
    \label{eq::probabilistic_dynamics}
\end{align}
where $D$ is any arbitrary probability distribution whose moments depend on $x,u$, provide better representation of real-world scenarios. Integrating $x$ under~\eqref{eq::probabilistic_dynamics} is analytically intractable in general, and several numerical schemes have been introduced~\cite{deisenroth2011pilco,amadio2022model}. ECUT, standing for Expansion Compression Unscented Transform, introduced in~\cite{Hardik_TCST_2025_foresee}, uses the Unscented Transform (UT) based on deterministic sampling to approximate a state distribution and propagate it through the dynamics function. Under state-dependent disturbances, though, modeled frequently using Gaussian Process or Bayesian Neural Networks, each sample must be mapped to multiple particles to account for increased uncertainty, leading to poor scalability when predicting future states for multiple time steps. Our Expansion Compression (EC) UT scheme overcomes this issue by first increasing the number of samples as needed but then reducing them based on moment matching to arrive at best set of finite samples that lead to minimum loss of information. ECUT is used to predict future states efficiently for trajectory optimization and local planning methods like MPC~\cite{Hardik_TCST_2025_foresee} and MPPI~\cite{parwana2024risk}. It is also used in an online constrained policy optimization framework FORESEE, to tune controller parameters on the fly as the robot navigates the environment~\cite{Hardik_TCST_2025_foresee}.

\subsection{Safe and Robust Observer-Controller Synthesis Using Control Barrier Functions}\label{observer_controller}

The work introduced in~\cite{Dev_CSL_2023} addresses the challenge of ensuring safety in nonlinear systems with partial and noisy state measurements. Traditional safety-critical control methods, such as those based on Control Barrier Functions (CBFs), often assume perfect state knowledge, which is unrealistic due to sensor noise and external disturbances. This limitation makes it difficult to enforce safety constraints reliably when the true system state is uncertain.

To solve this, we proposed an observer-controller framework that integrates state estimation with CBF-based control. We develop two approaches: one using Input-to-State Stable (ISS) observers, which ensure bounded estimation errors dependent on disturbances, and another using Bounded Error observers, which provide deterministic error bounds. These observer designs account for state estimation uncertainty, which is then incorporated into the safety constraints enforced by the CBF-based controller. The resulting control strategy is implemented using a quadratic program (QP), ensuring computational efficiency and real-time applicability. 

\subsection{Robust Leader-Follower Formation Control for Human-Robot Scenarios}\label{robust_leader_follower}

The work in~\cite{Alia_ACC_2022} extends previous control frameworks for leader-follower formation to ensure safety and robustness in human-robot scenarios. The goal is to maintain a safe, connected formation of aerial robots (followers) around a moving human (leader), under uncertainty in both the robots’ and the leader’s state estimates. The controller is based on Lyapunov-like barrier functions for safety, connectivity, and convergence. Uncertainty about the robots' states is quantified via a Kalman filter, while an adaptive active-passive estimator provides the human’s position and velocity.Safety is enforced probabilistically by inflating inter-robot safety distances and shrinking the connectivity region based on estimated error bounds. The proposed architecture guarantees convergence to a neighborhood of the desired formation and enforces collision avoidance with high probability. The framework was validated in hardware experiments with a team of quadrotors tracking a human trajectory outdoors. 

\section{Demonstration Through Hardware Experiments - Introducing DevQuad}
All hardware experiments are summarized in \cref{tab:Experiments}, with snapshots in \cref{fig:snapshots}. Recent experiments used our newly developed in-house DevQuad.

DevQuad (Devansh's Quad), shown in \cref{fig:snapshots}, is a custom agile quadrotor designed to optimize payload capacity and maximize flight time. The quadrotor has a wet weight of 0.820~kg and offers a 15-minute hover flight time. It is equipped with an onboard NVIDIA Orin NX, providing GPU-accelerated compute for real-time planning, perception, and control. The low-level geometric tracking controller is implemented on a Pix32V6c, which communicates with the Orin over UART. The main sensor payload is the Intel Realsense D455. DevQuad is open-source and open-hardware, with all software required to run the quad released, including the low-level PX4-Autopilot firmware with ground station support. Instructions for building the hardware are also publicly available. \href{https://dasc-lab.github.io/robot-framework/vision_drone/vision_drone_guide.html}{[https://dasc-lab.github.io/robot-framework/vision\_drone/vision\_drone\_guide.html]}

\nocite{*}
\bibliographystyle{IEEEtran}
\bibliography{ICRA_2025_Ws/main.bib}


\end{document}